# SMS Spam Detection and Classification to Combat Abuse in Telephone Networks Using Natural Language Processing

## Dare Azeez Oyeyemi [a] and Adebola K. Ojo [a*]

[a] *Computer Science Department, University of Ibadan, Nigeria.*

*Authors' contributions*

*This work was carried out in collaboration between both authors. Both authors read and approved the final manuscript.*

*Original Research Article*

## Abstract

In the modern era, mobile phones have become ubiquitous, and Short Message Service (SMS) has grown to become a multi-million-dollar service due to the widespread adoption of mobile devices and the millions of people who use SMS daily. However, SMS spam has also become a pervasive problem that endangers users' privacy and security through phishing and fraud. Despite numerous spam filtering techniques, there is still a need for a more effective solution to address this problem [1]. This research addresses the pervasive issue of SMS spam, which poses threats to users' privacy and security. Despite existing spam filtering techniques, the high false-positive rate persists as a challenge. The study introduces a novel approach utilizing Natural Language Processing (NLP) and machine learning models, particularly BERT (Bidirectional Encoder Representations from Transformers), for SMS spam detection and classification. Data preprocessing techniques, such as stop word removal and tokenization, are applied, along with feature extraction using BERT.
Machine learning models, including SVM, Logistic Regression, Naive Bayes, Gradient Boosting, and Random Forest, are integrated with BERT for differentiating spam from ham messages. Evaluation results revealed that the Naïve Bayes classifier + BERT model achieves the highest accuracy at 97.31% with the fastest execution time of 0.3 seconds on the test dataset. This approach demonstrates a notable enhancement in spam detection efficiency and a low false-positive rate.

___________

*\*Corresponding author: Email: adebolak.ojo@gmail.com;*

The developed model presents a valuable solution to combat SMS spam, ensuring faster and more accurate detection. This model not only safeguards users' privacy but also assists network providers in effectively identifying and blocking SMS spam messages.

# 1 Introduction

Mobile phones are now considered to be a kind of devoted companion to users. The widespread adoption of mobile phones, particularly for SMS communication, has become an integral part of modern life. Short Message Service (SMS) is a valuable service offered by the telecommunications industry and it contributes significantly to the Gross National Income (GNI) of developing countries [2]. This facility is used by millions of users daily due to its simplicity, accessibility, instant delivery, and low price rates as compared to calls. However, the ubiquity of SMS has also led to an increase in unwanted spam messages, including advertisements and scams. Nigeria, in particular, faces a significant SMS spam problem which has endangered mobile users' privacy with phishing and fraud [3]. Nigeria is ranked 3rd among the top 10 African countries affected by SMS spam according to a report by Truecaller. According to the report, an average mobile user in Nigeria received 35 spam messages per month. Fig. 1 shows the statistics of the category of Top SMS spammers in Nigeria and Fig. 2 shows the statistics of Average SMS spam received by Truecaller users in Africa.

Spam detection has traditionally relied on keyword filters to differentiate between spam and legitimate messages for the past two decades [4]. More recently, advanced methods like Statistical Learning Theory, Artificial Neural Networks (ANNs), and Support Vector Machines (SVM) have emerged. However, according to [5] these newer techniques exhibit inconsistent performance across different training datasets without logical or apparent explanation. There are numerous spam filtering techniques, however, because each of these techniques has strengths and drawbacks, no single spam filtering strategy can be guaranteed to be 100% effective at eradicating spam issues. The application of text mining techniques to SMS will improve the effectiveness of detecting and classifying spam messages [6], which will reduce telephone network abuse. There are now a staggering number of different forms of SMS, and a different technique that can effectively classify SMS with low latency must be proposed.

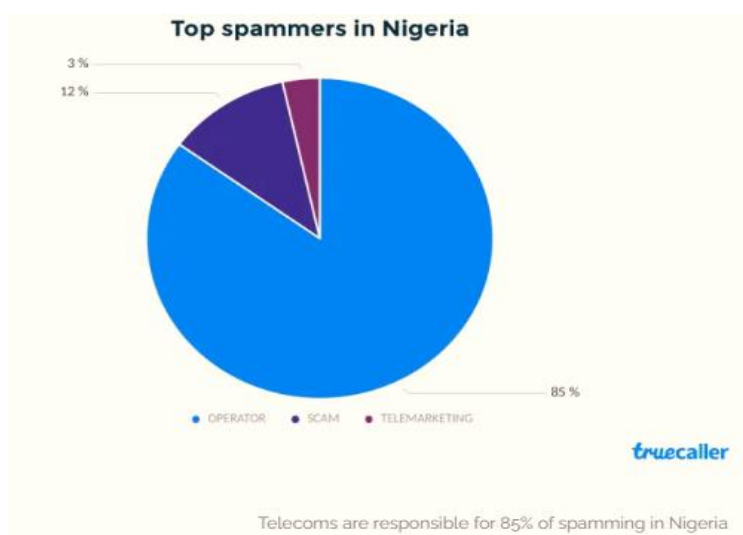

**Fig. 1. Category and Percentage (%) of Top SMS Spammers in Nigeria**
**(www.technext24.com/2019/12/04/truecaller-report)**

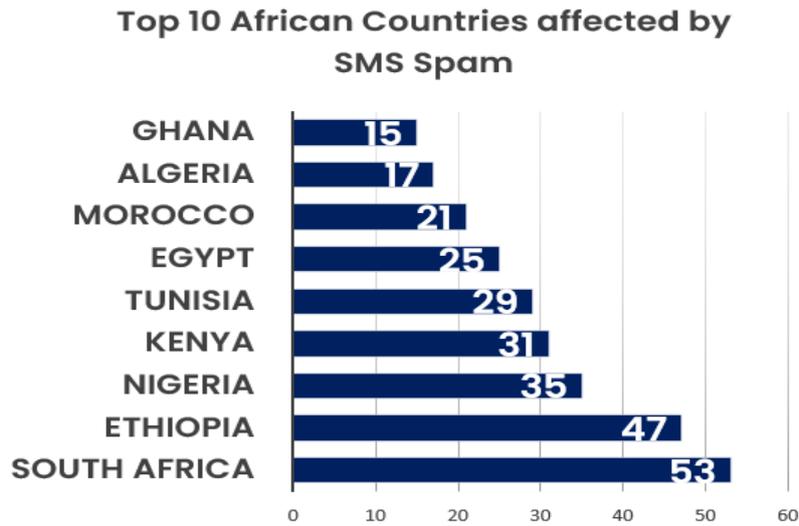

**Fig. 2. Truecaller Average spam SMS per user/month in Africa (https://pmnewsnigeria.com/1029/03_02)**

This study proposed the use of contextualized embedding, an NLP approach, to enhance SMS spam detection and classification. Contextualized embedding represents words based not only on their static definitions but also on the context in which they appear. By applying BERT to create contextual sentence embeddings for SMS messages, combined with machine learning algorithms like Naive Bayes, Random Forest, Gradient Boosting, Logical Regression, and SVM, this research aims to improve efficiency and accuracy in identifying spam messages in the lowest possible time.

## 2 Problem Statement

SMS is a text messaging platform that allows users to exchange short text messages using standardized communication protocols, [7]. These messages are succinct, filled with abbreviations, and less formal, posing a challenge in spam classification. Service providers show less concern about SMS spam due to its low occurrence and their use of SMS for promotions.

In Nigeria, there's an existing solution called Do Not Disturb (DND) offered by network providers through the Nigeria Communication Commission, but it's not entirely effective because it can also block legitimate messages [3]. Moreover, the unavailability of large SMS spam datasets used for training and testing spam detection models has led to issues of overgeneralization and overfitting, reducing the accuracy of classification predictions.

Given the limited character length of SMS and the evolving strategies employed by spammers, a key question emerges: Are the current models sufficiently effective and efficient in quickly distinguishing between SMS spam and legitimate messages? Addressing this question is pivotal in enhancing the accuracy and efficiency of SMS spam detection and classification.

## 3 Research Objectives

1. To pre-process the dataset using Natural Language Processing techniques using the BERT (Bidirectional Encoder Representations from Transformers) model.
2. To perform feature extraction and selection using Document Frequency Matrix and BERT preprocessor.
3. To perform vectorization to convert the processed dataset to numerical data.
4. To evaluate and compare the outcomes of the utilized machine learning models, which are Naive Bayes, Random Forest, Gradient Boosting, Logical Regression, and SVM.

# 4 Scope and Limitation(s) of Study

The study aims to develop an effective SMS spam detection and classification that leverages machine learning algorithms to combat abuse in the telephone network. It involves gathering a diverse dataset of SMS messages from various sources such as Kaggle, UCI, Data Science Nigeria, and self-sourced data, preprocessing the datasets it to remove noise and extract relevant features. These features encompass text-based and metadata-based attributes.

To identify spam or legitimate messages, this study employs machine learning algorithms, including fine-tuning the pre-trained BERT model for the SMS dataset to improve its contextual understanding of sentences of the SMS domain and generate highly relevant contextualized sentence embeddings. Machine language algorithms such as Naive Bayes, Random Forest, Gradient Boosting, Logical Regression, and Support Vector Machine are used, and their performances are compared using evaluation metrics.

One challenge is the dataset's class imbalance, with far fewer spam records than legitimate ones, requiring a down-sampling technique to the ham class, ensuring parity between ham and spam records. The limited dataset for training due to the down-sampling poses a potential performance challenge and it might affect model training as the dataset size increases. Additionally, the study is limited to the English language due to dataset composition, even though it can identify some pidgin words. The models may not effectively recognize slang or abbreviations, potentially affecting results when non-English words are used.

# 5 Literature Review

## 5.1 Related works

There have been several studies about SMS spam detection and classification in recent years. Researchers have proposed various techniques ranging from machine learning to hybrid approaches to detect and classify spam messages. Keyword Filtering is the most common approach used in SMS spam according to [8] but this approach has limitations, as spammers easily modify words they use or intentionally misspell to evade detection.

**5.1.1 Machine learning approaches**

In the study conducted by [9], they tested several supervised machine learning models including an unsupervised model K-means, Decision Trees, Naïve Bayes, SVM, and K-Nearest Neighbor on the UCI Machine Learning Repository dataset. They then developed a hybrid model that combines unsupervised and supervised algorithms to classify SMS messages as ham or spam. Among the combinations tested (Kmeans-SVM, Kmeans-NB, and Kmeans-LR), the Kmeans-SVM combination achieved the best accuracy of 98.8%. They did not explore other combinations of unsupervised and supervised algorithms to create a hybrid model.

[10] proposed a message topic model for spam SMS detection in 2021. Utilizing the KNN algorithm to address message sparsity, the model considered symbol terms and background terms, outperforming the standard LDA model. [10] undertook a similar investigation employing two datasets: one sourced from UCI machine learning, mirroring Kaggle's corpus, featuring 5,574 spam and ham messages; the other encompassing 2,000 messages of both types. Leveraging TF-IDF matrices, a comprehensive suite of machine learning algorithms, including CNN, Naive Bayes, SVM, Random Forest, ANN, and Decision Tree, were applied to the two datasets. CNN achieved state-of-the-art results with 99.10% accuracy.

From the above research, it becomes evident that traditional algorithms like Naive Bayes and SVM consistently outperform other methods. Moreover, considering message length as an additional feature could potentially contribute to the model's overall performance.

**5.1.2 Deep learning approaches**

[11] adopted a deep learning approach called BiLSTM (Bidirectional Long Short-Term Memory), which is a type of Recurrent Neural Network (RNN), for detecting spam messages. The outcome of their study indicates

that this model performed better than alternative machine learning algorithms, including BayesNet, J48, Naïve Bayes, SVM, K-nearest neighbor, and decision tree. Notably, the model attains an accuracy rate of 98.6%. They used the Word2vec algorithm for word embedding. However, the model had some limitations such as high preprocessing time due to the manual removal of unstandardized abbreviations. Deep learning techniques have shown promising results in SMS spam detection and classification. [12] titled their paper, "Deep Learning to Filter SMS Spam". They utilized Convolutional Neural Network (CNN) and Long Short-Term Memory (LSTM), the models were based solely on text data only and achieved a high accuracy of 99.44%.

### 5.1.3 Ensemble learning and Feature engineering approaches

Ensemble learning involves combining multiple models to improve accuracy and reduce overfitting. These techniques have also been applied to the domain of SMS spam detection and classification. [13] titled their research "Enhancing Spam Message Classification and Detection Using Transformer-Based Embedding and Ensemble Learning". In this work, the authors used an ensemble learning approach for SMS spam detection by combining four machine-learning models into one model. The model performed better than its separate constituent parts achieving a high accuracy of 99.91%.

Feature engineering techniques have been applied to improve the performance of SMS spam detection and classification models. [14] introduced an approach rooted in feature engineering, leveraging semantic analysis to extract features from SMS messages. They created a dictionary using the TF-IDF Vectorizer algorithm, which includes all the features of words of a spam SMS. The system classifies SMS by referring to this dictionary and based on the message content.

# 6 Methodology

## 6.1 Research Design and Approach

The research design and approach for this study used a cross-sectional research method that involves collecting and analyzing data at a specific time. The research design is structured into: data collection, data preprocessing, feature extraction, and classification. The study aims to develop a model for SMS classification and detection that can be used to combat abuse in a telephone network.

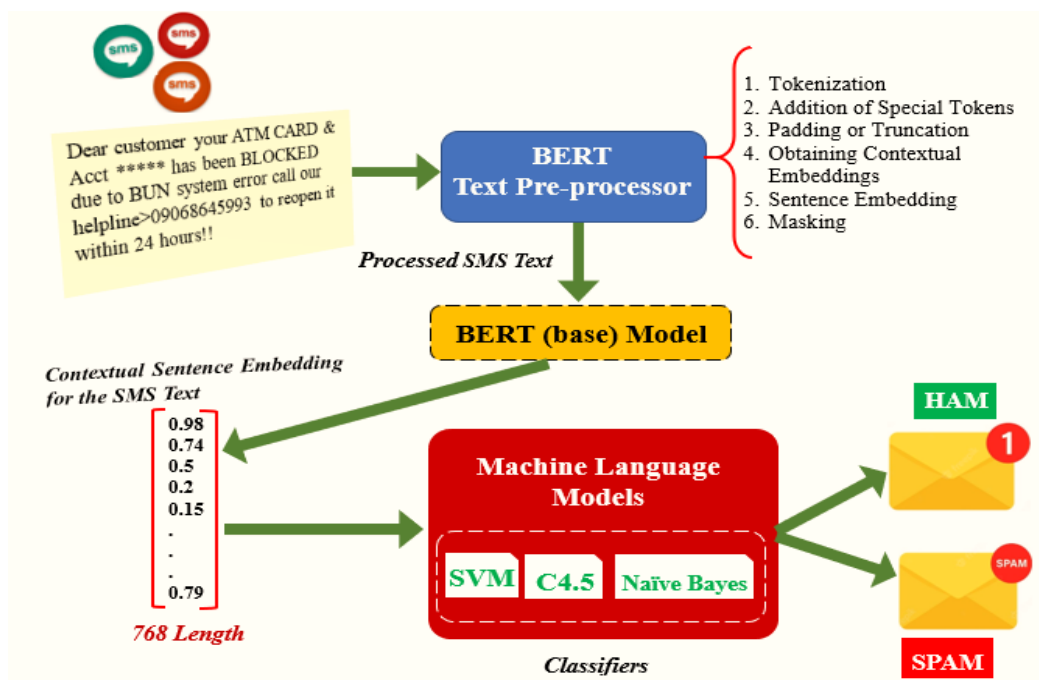

**Fig. 3. Proposed Methodology for the study**

**6.1.1 Cross-sectional study**

For this research, a cross-sectional study was used to collect data on SMS messages from various online repositories. A large number of mobile users in Nigeria fell victim to SMS spam during the COVID-19 period (2020/21) compared to SMS spam data gathered from previous years [15]. The spam messages data collected represent a snapshot of the SMS messages at that specific point in time, which was used to detect and classify spam messages in the telephone network. The advantage of a cross-sectional study is that it makes it possible for the collection of data from a large population sample at a specific point in time. This approach allows the testing of hypotheses and the drawing of statistical inferences, providing insights into the effectiveness of SMS spam detection and classification systems.

**6.1.2 Data collection phase**

At this primitive level, data is gathered locally and globally from numerous sources to create a respectable dataset of spam and ham text messages, which will be utilized as the model's input (SMS messages). The spam dataset used for this research contained a total of 6,986 rows.

1. Kaggle: The dataset contains 5,572 rows of SMS classified into ham (nonspam) and spam.
2. Data Science Nigeria (DSN): the dataset contains 1,141 text messages. The dataset contains fraudulent messages in the financial and labor sectors received in any location in the country.
3. Self-Data: Google form to collect local spam messages mobile users received (275 spam messages)

The main reason for choosing the dataset is to combine them by randomly sampling from different resources to give exposure to different scenarios of non-spam and spam SMSs.

**6.1.3 Data cleaning and preprocessing**

The three datasets were cleaned to have uniform classes and data before we performed preprocessing on the SMS spam dataset utilizing Python and its libraries such as NLTK library and others. Preprocessing is the initial stage of turning the collected dataset (unstructured data) into more structured data and converting the input text into numerical representations that can be used for various downstream natural language processing (NLP) tasks. For the SMS Spam dataset used in this study, preprocessing was done to remove unnecessary elements that don't contribute to the classification task such as stop words, special characters, and punctuations before applying the classification models. Stemming was also performed to minimize the word count within the messages and converting the text into lowercase to avoid case sensitivity issues, while still preserving their meaning.

BERT model preprocessor and encoder aided the conversion of the dataset into a numerical format used for the various downstream natural language processing tasks in this study. The preprocessor prepared the input data for the BERT model by performing tokenization, adding special tokens such as [CLS] and [SEP], segmenting the input data into separate sequences, masking tokens, and padding the input sequence to a fixed length. Fig/ 4 and 5 shows how the pre-trained BERT model process text data.

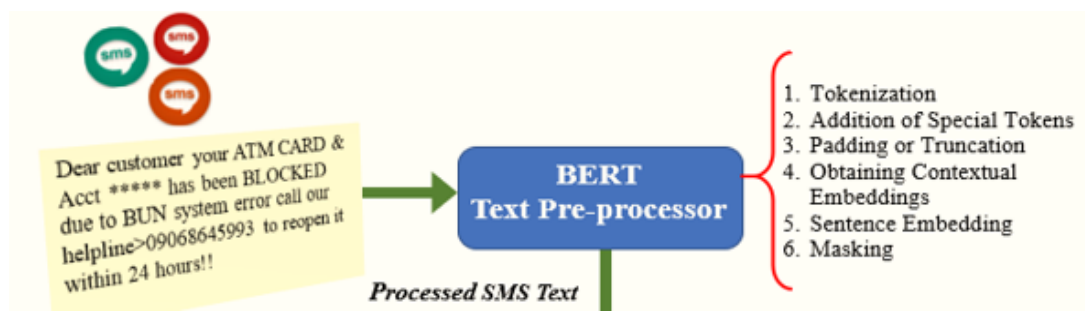

**Fig. 4. BERT model text pre-processor**

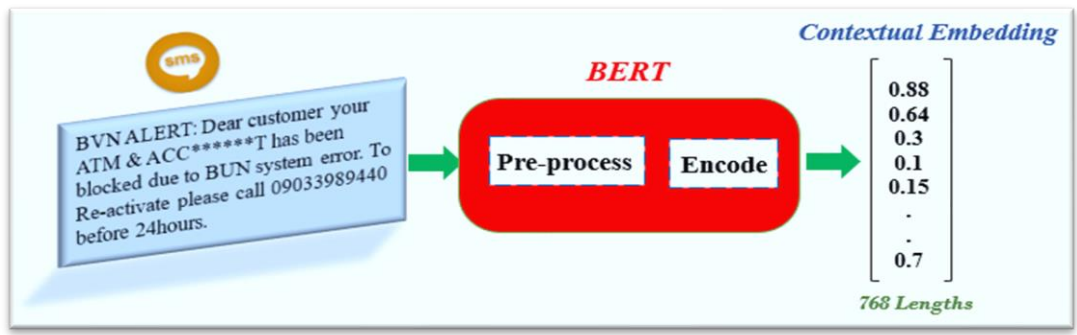

**Fig. 5. Conversion of Text Data to Contextual Embedding in BERT Model**

The encoder processes the preprocessed data to generate contextualized word embeddings. The encoder is made up of multiple transformer layers, that capture the contextual relationships between the words in the input text. The contextualized embeddings are then passed to the next transformer layer until the final layer generates a fixed-size vector representation for the input data.

**6.1.4 Feature extraction and selection**

Feature extraction and selection are important to balance the overfitting and overgeneralization of ham and spam in SMS message classification. The features extracted from the preprocessed data were done using both text-based and metadata-based approaches. The text-based features include word frequency and part-of-speech tags, which capture the linguistic characteristics of spam messages while the metadata-based features include sender information and time of message, which provide additional information about the likelihood of a message being spam.

To represent the SMS messages as numerical vectors for classification, the pre-trained BERT model was utilized to extract contextualized word embeddings for each token in the tokenized SMS messages. BERT generates these embeddings by considering the entire sentence's context, capturing the nuanced meaning of words based on their surroundings. These embeddings contain rich semantic and syntactic information, enabling better representation of words in the SMS messages. The dataset was split into 4540 rows for training, 1396 rows for testing, and 1050 rows for validation. The training dataset was used to train the BERT model for the spam detection task, involving the fine-tuning of model parameters and weights. Subword tokenization was performed by the pre-trained BERT model to handle vocabulary variability and capture the meaning of rare or complex words in the messages, especially Pidgin English and other local languages. The test dataset was used for model evaluation.

**6.1.5 Classification**

To classify SMS messages as either ham or spam, both rule-based filtering and machine learning-based classification methods were utilized. The rule-based filter was applied to the preprocessed data to quickly discard obvious spam messages. This filter was based on a set of rules defined using the characteristics of known spam messages while the remaining messages were classified using machine learning-based classification. During this phase, machine learning classifiers such as Logistic Regression, Naive Bayes, Random Forest, Gradient Boosting, and SVM in conjunction with the BERT model were utilized.

# 7 Results and Discussion

## 7.1 Rule-based filtering approach

In this study, rule-based filtering was employed to automatically detect and filter out spam SMS messages. This approach was based on a set of predefined rules and known characteristics of spam messages such as win free money, job interviews, and BVN (Bank Verification Number) updates. The rules included keywords and phrases commonly associated with spam messages. The predefined rules were categorized as follows:

1. BVN Spam: SMS messages containing phrases or keywords like "BVN ALERT", "BVN system error", "Dear Customer your ATM card and Account has been Blocked Due to BVN error", "CENTRAL BANK has blocked your Account and Atm Card has been deactivated", and more.
2. Investment Spam: SMS messages containing shortened links and keywords or phrases such as "your account has been credited with $20,000 register within 24hrs to claim it", "We need to validate your account click http://bit.ly/2lh54", "Security Watch", "http://goo.gl/a93k", and more.
3. Fake Job Spam: Messages containing keywords or phrases like "You are invited for an aptitude test with UBA Plc", "MAG CONSULT", "MATRIXGLOVER", "invited for an interview at FOSAD CONSULTING LIMITED", "Congratulations!!! You are invited to take part in an interview session at M.H.S", "You have been selected to work at SHELL Oil in Lagos, call Mr. John" and more.
4. Marketing or Advertisement Spam: SMS messages containing phrases or keywords such as "win", "AWOOF CASH", "lottery", "Enjoy More financial freedom", "Get double your cash" and more.

The system evaluates incoming SMS message to determine if they matched any of the predefined rules. If a message match one or more of the above predefined rules, it is flag as spam; otherwise, it is considered ham or non-spam message. While the rule-based filtering approach is effective at identifying and classifying obvious spam messages with known characteristics, it may not capture all types of spam, especially those employing sophisticated or evolving tactics. To enhance accuracy, the remaining messages were classified using machine learning-based techniques.

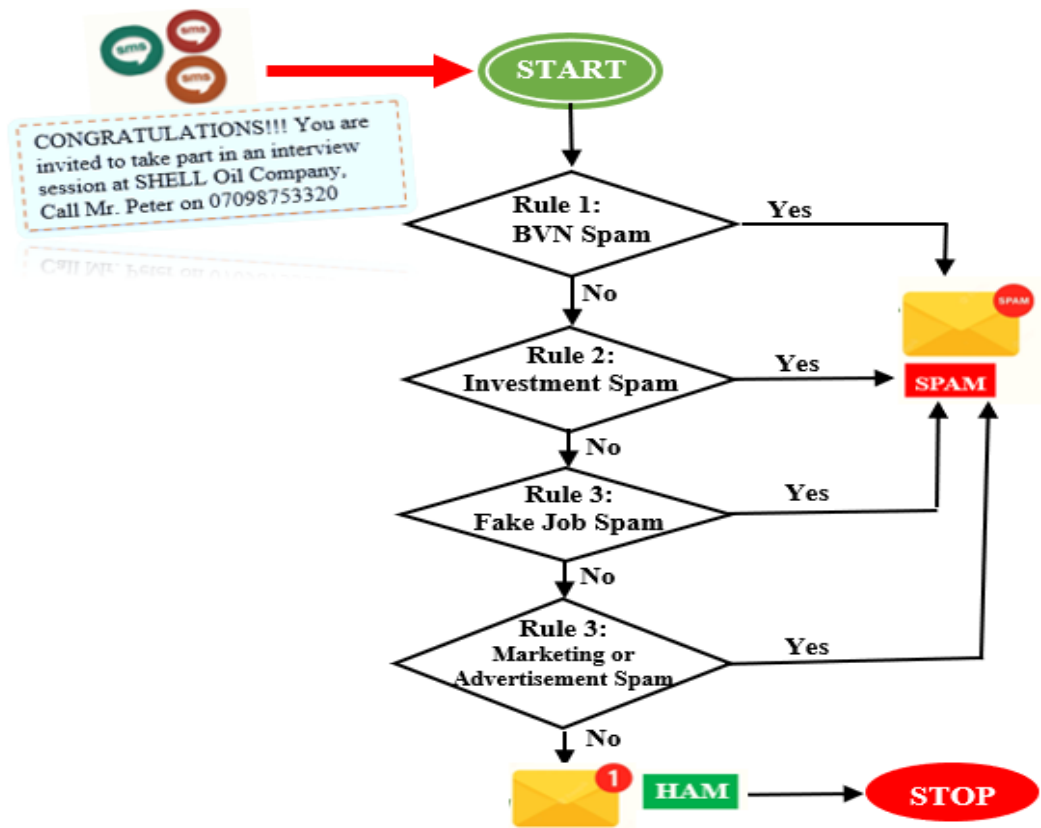

**Fig. 6. Flowchart of Rule-Based Approach to SMS Spam**

## 7.2 Vectorization

To run machine learning algorithms, text files must be converted into numerical feature vectors, this process is known as Vectorization [16]. Machine learning models operate exclusively on numerical data, to achieve this, a matrix was created to contain words and their frequency of occurrence. The two techniques used for the creation of the document term matrix in this study are Bag of Words (BoW) and Term frequency-inverse document frequency (TF-IDF).

1. Bag of words (BoW): Bag of Words is a way of extracting the features from the set of text messages [14]. Fig. 7 shows a matrix called bagOfWords which describes the text based on the frequency of the word appearing in the document and was implemented using Python's package CountVectorizer.
2. Term frequency-inverse document frequency (TF-IDF): The TF-IDF matrix helps in understanding the importance of the word in the corpus of documents [14]. To implement, the TF-IDFVectorizer() function was used with the n-gram as a unigram. Fig. 8 shows the matrix of the converted dataset to numerical data used for the model. The machine learning models were implemented after the creation of the matrix.

```
   Text_Length  1  2  3  4  5  6  ...  3615  3616  3617  3618  3619  3620  3621
0           29  0  0  0  0  0  0  ...     0     0     0     0     0     0     0
1          101  0  0  0  0  0  0  ...     0     1     0     0     0     0     0
2          141  0  0  0  0  0  0  ...     0     0     0     0     0     0     0
3           59  0  0  0  0  0  0  ...     0     0     0     0     0     0     0
4           31  0  0  0  0  0  0  ...     0     0     0     0     0     0     0
```

**Fig. 7. Generated Bag of Words Matrix for Spam dataset**

```
   Text_Length    1    2    3    4    5  ...      3616  3617  3618  3619  3620  3621
0         29.0  0.0  0.0  0.0  0.0  0.0  ...  0.000000   0.0   0.0   0.0   0.0   0.0
1        101.0  0.0  0.0  0.0  0.0  0.0  ...  0.286945   0.0   0.0   0.0   0.0   0.0
2        141.0  0.0  0.0  0.0  0.0  0.0  ...  0.000000   0.0   0.0   0.0   0.0   0.0
3         59.0  0.0  0.0  0.0  0.0  0.0  ...  0.000000   0.0   0.0   0.0   0.0   0.0
4         31.0  0.0  0.0  0.0  0.0  0.0  ...  0.000000   0.0   0.0   0.0   0.0   0.0
```

**Fig. 8. Generated TF-IDF Matrix for Spam dataset**

### 7.3 Comparing research result

It is essential to strike a balance between minimizing false positives (to avoid inconveniencing users) and false negatives in SMS spam classification (to catch all spam) is essential. Table 1 shows the comparison of the results of machine learning models used in this study.

From Table 1, the best-performed model in terms of accuracy, low latency, precision, and recall is the Naive Bayes + BERT model. The model has an accuracy of 97.31%, precision of 98%, recall of 99%, epoch time of 1 second on the training dataset, and inference time of 0.3 seconds on the test dataset. The model also has the lowest total count of False-Positive (FP) and False-Negative (FN): 30 compared to other models used in the study.

### 7.4 Confusion matrix of the models

Confusion matrix visualizes and summarizes the performance of a classification algorithm [13]. In classification problems, reviewing the confusion matrices of the models used is crucial for assessing their performance. Table 2 shows the confusion matrices for all the models in this study.

### 7.5 Comparison between developed and existing models

Table 3 shows the comparison of the existing models and the proposed model.

**Table 1. Comparison of utilized Models**

|   | ML Classifiers | Training Accuracy | Training Execution Time | Testing Accuracy | Testing Execution Time | Recall | Precision | FP | FN |
|---|---|---|---|---|---|---|---|---|---|
| 1 | SVM + BERT | 93.78% | 8 seconds | 94.35% | 1 seconds | 95% | 98% | 47 | 16 |
| 2 | Logical Regression + BERT | 95.29% | 2 seconds | 96.14% | 0.8 seconds | 97% | 99% | 29 | 14 |
| 3 | Gradient Boosting + BERT | 95.75% | 3 minutes 9 seconds | 96.5% | 1 minute 28 seconds | 98% | 98% | 22 | 17 |
| **4** | **Naive Bayes + BERT** | **96.83%** | **1 seconds** | **97.31%** | **0.3 seconds** | **99%** | **98%** | **13** | **17** |
| 5 | Random Forest + BERT | 92.31% | 16 minutes 11 seconds | 92.55% | 14 minutes 4 seconds | 93% | 98% | 63 | 20 |

**Table 2. Confusion matrix of utilized Models**

|   | ML Classifiers | Confusion Matrix | Predicted Spam SMS Message | Predicted Non-Spam SMS Message |
|---|---|---|---|---|
| 1 | SVM + BERT | Truly Spam | 255 | 16 |
|   |   | Truly Non-Spam | 47 | 1078 |
| 2 | Logical Regression + BERT | Truly Spam | 257 | 14 |
|   |   | Truly Non-Spam | 29 | 1096 |
| 3 | Gradient Boosting + BERT | Truly Spam | 254 | 17 |
|   |   | Truly Non-Spam | 22 | 1103 |
| 4 | **Naive Bayes + BERT** | **Truly Spam** | **261** | **13** |
|   |   | **Truly Non-Spam** | **17** | **1105** |
| 5 | Random Forest + BERT | Truly Spam | 251 | 20 |
|   |   | Truly Non-Spam | 63 | 1062 |

**Table 3. Comparison of existing models with proposed model**

| References | Method | Model | Accuracy | Comments |
|---|---|---|---|---|
| S. Gupta, Saha, and Das [17] | TF-IDF vectorization algorithm | Did not state the specific machine learning model utilized | 96.5% | The authors employed and assessed using only one vectorization technique. |
| Adebayo Abayomi-Alli, Olusola, and Sanjay Misra, [11]. | Word Embedding (Word2vec Algorithm) | BiLSTM (Bidirectional Long Short-Term Memory) | 98.6% | Very high preprocessing time |
| Baaqeel, Hind, and Rachid Zagrouba [9] | Word Tokenization | Kmeans-SVM | 98.8% | Did not investigate alternative combinations of unsupervised and supervised algorithms for building the hybrid model. |
| Current Research | Contextual Sentence Embedding | Naive Bayes + BERT | 97.3% | Class imbalance issue rectified, detect local languages (Pidgin), less execution time |

# 8 Conclusion

This study aims to design a proficient machine learning model capable of identifying spam SMS, thus addressing the issue of telephone network abuse prevalent in Nigeria. In addition to this primary goal, a supplementary contribution of this research involves juxtaposing the outcomes of fully developed models against existing counterparts using evaluation metrics and visual representations. These dual contributions are poised to significantly benefit mobile phone users by expeditiously mitigating the challenge of spam SMS, consequently streamlining its detection process. A natural language processing approach was used to achieve this objective, with a particular emphasis on understanding the semantics of the text. In this study, a hybrid model, Naïve Bayes, and BERT model achieved an impressive accuracy rate of 97.3%, a precision of 98%, and a recall of 99% with an execution time of 0.3 seconds. All these indicate that the model is highly effective in identifying ham messages correctly while also being able to distinguish spam messages in less time. Looking ahead, extending the scope to encompass non-English languages presents an intriguing prospect for advancing spam SMS detection in upcoming research.

Looking ahead, there is significant potential for this study to be expanded. In the future, we can focus on collecting new latest data and testing the proposed model. To enhance the efficacy of the results, an avenue for improvement involves training machine learning models using localized datasets sourced from diverse countries, while also considering datasets with substantial records. This approach holds the potential to bolster the dependability of the predictive models. Furthermore, extending the scope to encompass non-English languages presents an intriguing prospect for advancing spam SMS detection in upcoming research endeavors.

## Competing Interests

Authors have declared that no competing interests exist.

## References

[1] Mageshkumar N, Vijayaraj A, Arunpriya N, Sangeetha A. Efficient spam filtering through intelligent text modification detection using machine learning, Mater. Today Proc. 2022;64:848–858.
DOI: https://doi.org/10.1016/j.matpr.2022.05.364

[2] Sjarif NNA, Azmi NFM, Chuprat S, Sarkan HM, Yahya Y, Sam SM. SMS Spam Message Detection using Term Frequency-Inverse Document Frequency and Random Forest Algorithm, Procedia Comput. Sci; 2019, [Online].
Available:https://api.semanticscholar.org/CorpusID:213854042

[3] Marcus A. Effect of SMS Advertising on Attitudes of Nigeria GSM Phone Users, Apr. 2019;3:215–225.

[4] Hussain N, Mirza H, Hussain I. Detecting Spam Review through Spammer's Behavior Analysis, ADCAIJ Adv. Distrib. Comput. Artif. Intell. J. Mar. 2019;8:61.
DOI: 10.14201/ADCAIJ2019826171

[5] Saraswathi D, Sowmya D. SMS Spam Classification Using PSO-C4.5, Lect. Notes Electr. Eng. 2023;967:41–47.
DOI: 10.1007/978-981-19-7169-3_4

[6] Ranjith Reddy K, Chaudhary S. An Efficient Text Mining Technique and Its Application to SMS Spam Detection BT - Data Engineering and Intelligent Computing, V. Bhateja, L. Khin Wee, J. C.-W. Lin, S. C. Satapathy, and T. M. Rajesh, Eds., Singapore: Springer Nature Singapore/ 2022;201–213.

[7] Maruf A, Numan A, Haque M, Tahmida Jidney T, Aung Z. Ensemble Approach to Classify Spam SMS from Bengali Text; 2023.
DOI: 10.1007/978-3-031-37940-6_36

[8] Hanif EI, et al. Malay SMS Spam Detection Tool Using Keyword Filtering Technique, J. Phys. Conf. Ser. 2021;1793(1).
DOI: 10.1088/1742-6596/1793/1/012064

[9] Baqeel H, Zagrouba R. Hybrid SMS Spam Filtering System Using Machine Learning Techniques; 2020.
DOI: 10.1109/ACIT50332.2020.9300071

[10] Gupta S, Saha S, Das SK. SMS Spam Detection Using Machine Learning, J. Phys. Conf. Ser. Feb. 2021;1797:12017.
DOI: 10.1088/1742-6596/1797/1/012017

[11] Abayomi-Alli O, Misra S, Abayomi-Alli A. A deep learning method for automatic SMS spam classification: Performance of learning algorithms on the indigenous dataset, Concurr. Comput. Pract. Exp. 2022;34(17):e6989.
DOI: https://doi.org/10.1002/cpe.6989

[12] Roy PK, Singh JP, Banerjee S. Deep learning to filter SMS Spam, Futur. Gener. Comput. Syst. 2020;102:524–533.
DOI: https://doi.org/10.1016/j.future.2019.09.001

[13] Ghourabi A, Alohaly M. Enhancing Spam Message Classification and Detection Using Transformer-Based Embedding and Ensemble Learning. Sensors (Basel). 2023;23(8).
DOI: 10.3390/s23083861

[14] Das Gupta S, Saha S, Das SK. SMS spam detection using machine learning, J. Phys. Conf. Ser. 2021;1797(1).
DOI: 10.1088/1742-6596/1797/1/012017

[15] Godson N, Inyama HC, Ohaneme CO, Ozioko FE. The Impact of Using SMS Alert-Based E-Health in Increasing Outpatients Access to Health Care Resources in Nigeria, IOSR J. Comput. Eng. 2023;18:23–29.
DOI: 10.9790/0661-1806032329

[16] Rubin Julis M, Alagesan S. Spam detection in SMS using machine learning through text mining, Int. J. Sci. Technol. Res. 2020;9(2):498–503.

[17] Ballı S, Karasoy O. Development of content-based SMS classification application by using Word2Vec-based feature extraction, IET Softw. 2019;13(4):295–304.
DOI: 10.1049/iet-sen.2018.5046